\documentclass[runningheads]{llncs}
\usepackage[T1]{fontenc}
\usepackage{graphicx}
\usepackage{cite}
\usepackage{amsmath,amssymb} % define this before the line numbering.
\usepackage{multirow}
\usepackage{booktabs}
\usepackage{makecell}
\usepackage{subfig}
\usepackage{url}
\usepackage[colorlinks, citecolor=green, linkcolor=red]{hyperref}

\usepackage{color}

\begin{document}
\title{LHDR: HDR Reconstruction for Legacy Content using a Lightweight DNN}
\author{Cheng Guo\inst{1,2}\orcidID{0000-0002-2660-2267} \and Xiuhua Jiang\inst{1,2}}
\authorrunning{C. Guo and X. Jiang}
\institute{State Key Laboratory of Media Convergence and Communication, Communication University of China \and
Peng Cheng Laboratory, Shenzhen, China \\
\email{\{guocheng, jiangxiuhua\}@cuc.edu.cn}}
\maketitle
\begin{abstract}
High dynamic range (HDR) image is widely-used in graphics and photography due to the rich information it contains. Recently the community has started using deep neural network (DNN) to reconstruct standard dynamic range (SDR) images into HDR. Albeit the superiority of current DNN-based methods, their application scenario is still limited: (1) heavy model impedes real-time processing, and (2) inapplicable to legacy SDR content with more degradation types. Therefore, we propose a lightweight DNN-based method trained to tackle legacy SDR. For better design, we reform the problem modeling and emphasize degradation model. Experiments show that our method reached appealing performance with minimal computational cost compared with others.

\keywords{High dynamic range  \and Legacy Content \and Degradation model.}
\end{abstract}
%
%===========================================================
\section{Introduction}

Image's dynamic range is defined as the ratio of maximum recorded luminance to the minimum. As name implies, high dynamic range (HDR) image is able to simultaneously envelop rich information in both bright and dark areas, making it indispensable in photography and image-based lighting\cite{ReinhardHDRBook}, etc. The common way to obtain an HDR image is fusing multiple standard dynamic range (SDR) images with different exposure, i.e. multi-exposure fusing (MEF)\cite{Debevec97}. And recently the community has begun to use deep neural network (DNN)\cite{Kalantari17MEHDR,Wu18MEHDR,Yan19MEHDR,Yan20MEHDR,Chen21MEHDR,NTIRE22MEHDR} to tackle motion and misalignment between different SDR exposures.

While most MEF HDR imaging method is intended to be integrated in camera pipeline for taking new HDR photo, there is a considerable amount of legacy SDR content containing unreproducible historical scenes to be applied in HDR application. Those legacy content has limited dynamic range due to the imperfection of old imaging pipeline, and most importantly, no multi-exposure counterpart to be directly fused into HDR. In this case, we could only manage to recover HDR content from a single SDR image, which is called inverse tone-mapping (ITM)\cite{Banterle06ITM} or single-image HDR reconstruction (SI-HDR)\cite{CheatHDR}.

Different from MEF HDR imaging where full dynamic range is already covered in multiple SDR input, SI-HDR is an ill-posed problem since a method is supposed to recover the lost information by the reduction of dynamic range, etc. Fortunately, DNN has been proven effective in other ill-posed low-level vision tasks, hence researchers have begun to involve it in SI-HDR\cite{Eilertsen17,Marnerides18,Liu20,Santos20,Chen21}. DNN-based SI-HDR could better infer lost information e.g. those by highlight saturation, since DNN is able to aggregate and process semantic context.

\begin{figure}[t]
	\centering
	\includegraphics[width=\linewidth]{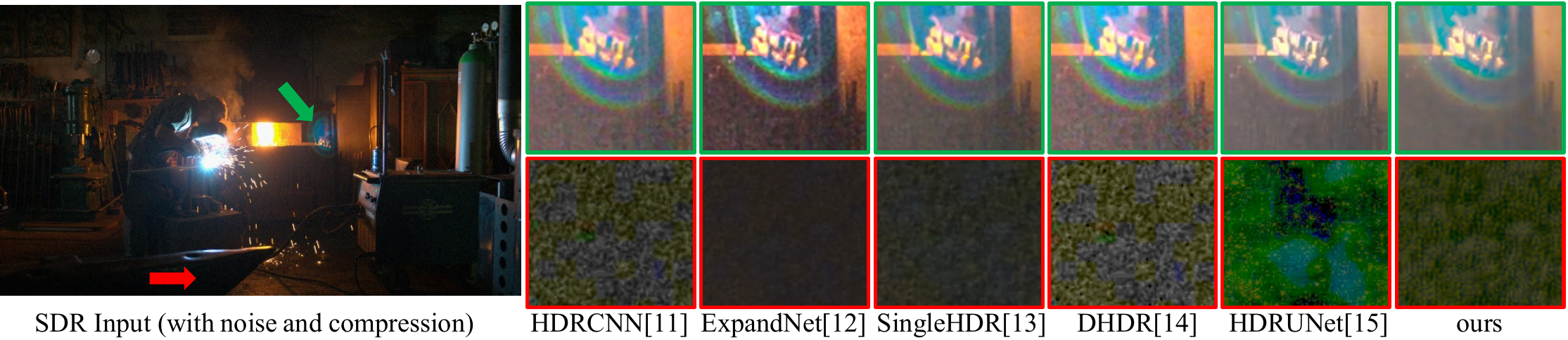} 
	\caption{Current HDR reconstrction methods struggle to handle SDR with noise and compression. In the green and red boxes are tone-mapped reconstrcted HDR.}
	\label{fig0}
\end{figure}

Albeit the success of current DNN-based SI-HDR, there are still 2 aspects to be considered: First, while legacy SDR content is susceptible to noise and compression, to the best of our knowledge, there is no DNN-based method motivated to jointly tackle both of them. Hence, current methods struggle to deal with legacy SDR with noise and compression, as Figure\ref{fig0} shows. Second, many method exploit a bulky DNN, which will hinder their real-time processing and deployment on devices with limited computational resources.

Therefore, our task is to design a lightweight DNN capable of handling legacy SDR with noise and compression. Our lightweight approach partly lies in pointwise and group convolution. Meanwhile, to teach the DNN with recovery ability, corresponding degradation should be correctly set in training. To this end, we clarify what kind of degradations are to be recovered by a reformed problem modeling based on camera pipeline\cite{BrownPipeline}. This problem modeling also helps us derive a DNN with modules customized to specific types of degradation. 

In the following paper, we first conclude related works from several aspects, then describe the DNN design containing problem modeling and network modules, and finally introduce our training strategy including degradations. Experiment will show that our method outperforms both DNN-based state-of-the-art (SOTA) and non-DNN method\cite{Rempel07EO} on both simulated and real legacy content. Finally, ablation studies are conducted to validate the effectiveness of our ideas.

Our contributions are:
\begin{itemize}
	\item To the best of our knowledge, making the first attempt handling legacy SDR content with both noise and compression in DNN-based SI-HDR.
	\item Lightening the DNN to facilitate its practical application.
	\item Reforming SI-HDR problem modeling by a precise camera pipeline, and emphasizing the impact of degradation model which has long been understated.
\end{itemize}

%===========================================================
\section{Related Work}

All following methods start from a single SDR image, yet they markedly differ regarding whether it directly outputs an HDR image.

\textbf{Direct approach.} The initiator of DNN-based SI-HDR \cite{Eilertsen17} reconstruct lost information in mask-split saturated region, and blends it with unsaturated area expanded by inverse camera response function (CRF). \cite{Marnerides18} first apply non-UNet DNN with multiple branches in different receptive field. \cite{Liu20} model SI-HDR with preliminary 3-step camera pipeline, and assign 4 sub-networks to hierarchically resolve them. \cite{Santos20} introduce more mechanisms: partial convolution from inpainting and gram matrix loss from style transfer. \cite{Chen21} append their UNet with an extra branch to involve input prior in spatial feature modulation. Ideas of other direct approach\cite{Zhang17,Jang18,Moriwaki18,Wang19,Soh19,Khan19,Marnerides21,Ye21,Lee21,Sharif21,Zhang21,Liu21,Raipurkar21,Yu21,Borrego22,Wu22} are omitted here, and will be detailed later if involved.

\textbf{Indirect approach.} Recovering lost information makes SI-HDR more challenging than MEF HDR imaging, hence some methods mitigate this difficulty indirectly. Common idea\cite{Endo17,Lee18,Lee18ECCV,Jo21,Banterle22} is to transfer single SDR image into multi-exposure SDR sequence to be later merged into HDR using traditional MEF algorithm. Other ideas include: learning the relationship between fixed and real degradation\cite{Kinoshita19}, non-I2I (image-to-image) histogram learning \cite{Jang20}, using spatial adaptive convolution whose per-pixel weight is predicted by DNN\cite{Cao21}, and polishing the result of 3 traditional ITM operators\cite{Chambe22}. We take direct approach since indirect one conflicts with the goal of efficient design.

%------------------------------------------------------------------------- 
\subsection{Lightweight SI-HDR}

Lightweight/efficient DNN has long received attention academically and industrially in other low-level vision tasks, e.g. super-resolution\cite{AIM20ESR,NTIRE22ESR}, denosing\cite{MobileAI21FDN}, and even MEF HDR imaging\cite{NTIRE22MEHDR}. Yet, few efforts are made in SI-HDR:

Preliminary attempts are made in \cite{Khan19} and \cite{Lee18ECCV} where feedback/recursive DNN module with shared parameters is used to reduce total number of DNN parameters (\#param), however, only \#param is reduced while computational cost is not. To be deployed on mobile platform, \cite{Borrego22} reduce computational cost by changing some parameter precision from \verb+float32+ to \verb+int8+. \cite{Wu22} apply a similar scheme as \cite{Eilertsen17} but use group convolution to lighten their DNN. \cite{Chambe22} is efficient mainly due to its pre/post-processing: their DNN only needs to polish single luminance channel resulting from existing methods.

Their efficiency is manifested in \#param, number of multiply-accumulate operations (MACs), and runtime, which will be detailed in experiment part.

%------------------------------------------------------------------------- 
\subsection{Problem Modeling of SI-HDR}

SI-HDR belongs to restoration problem, even using `black box' DNN, some methods still try to figure out exact degradations to restore. They assign specific degradation with a single step of sub-network and resolve them hierarchically.

Some methods divide their steps subjectively: \cite{Sharif21} into `denoising and exposure adjustment' and tone expansion; \cite{Liu21} into dequantization and hallucination; while that of \cite{Raipurkar21} are linearization, hallucination, and refinement.

Few are based on formulating a camera pipeline: \cite{CheatHDR} divide the task into inverting CRF, bit-depth expansion, and under\&over-exposure reconstruction, while steps of \cite{Liu20} are dequantization, linearization, hallucination, and refinement. All these step divisions or problem modeling could not cover both noise and compression from legacy SDR, our `target audience'.

%------------------------------------------------------------------------- 
\subsection{SI-HDR for Legacy Content}

Non-DNN method \cite{Rempel07EO} share the same hypothesis that legacy SDR content is susceptible to noise and quantization artifact, and uses filter to mitigate them. Though DNN has been widely studied in SI-HDR, denoising, and compression artifact removal separately, no DNN-based method is motivated to jointly handle all 3 tasks. Yet, there do exist methods tackling up to 2 of them simultaneously:

\textbf{Joint SI-HDR and compression artifact removal.} \cite{Eilertsen17} provide an alternative checkpoint trained with JPEG compressed SDR whose quality factor (QF) $\sim U(30,100)$ while \cite{Wang19} is trained with JPEG degradation with QF $\sim U(10,70)$.

\textbf{Joint SI-HDR and denoising.} Input SDR images in the training set of NTIRE HDR Challenge\cite{NTIRE22MEHDR,NTIRE21} contain zero-mean additive Gaussian camera noise, consequently all methods there will learn to jointly denoise. However, only \cite{Chen21,Sharif21} etc. belongs to SI-HDR, while the rest are MEF HDR imaging.

%------------------------------------------------------------------------- 
\subsection{HDR Degradation Model}
\label{sec:hdrdm}

From the above section, we know that only when specific degradation is added in training, the DNN will learn a corresponding restoration ability. Different from conventional image/video, there are exclusive steps in HDR-to-SDR degradation model due to their dynamic range discrepancy:

\begin{table}[b]
	\centering
	\caption{HDR-exclusive degradation models used by related works.}
	\label{tab1}
	\begin{tabular}{c|c|c|c|c} 
		\toprule
		\textbf{Type}                   & \textbf{Used by}     & \textbf{Nonlinearity}              & \textbf{Under-$\sim$} & \textbf{Over-exp. truncate}   \\ 
		\midrule
		\multirow{5}{*}{Simulated shot} & \makecell[c]{\cite{Eilertsen17,Santos20,Ye21,Zhang21} \\ \cite{Yu21,Wu22,Kinoshita19}} & \makecell[c]{virtual CRF \\ w. rand. param.} & $\times$ & \makecell[c]{{histogram fraction} \\ {$\sim U(5\%,15\%)$}}  \\
		\cline{2-5}
		& \cite{Liu20,Liu21,Endo17}     & rand. real CRFs           & \multicolumn{2}{c}{\multirow{2}{*}{rand. exposure adjust.}} \\ 
		\cline{2-3}
		& \cite{Raipurkar21}   & fixed CRF                          & \multicolumn{2}{l}{} \\ 
		\cline{2-5}
		& \cite{Wang19}        & `virtual cam.'                     & $\times$    & value $\sim U(0,10\%)$  \\ 
		\cline{2-5}
		& \cite{Chen21,Sharif21} etc.  & gamma$2.2\pm0.7$               & \multicolumn{2}{c}{fixed exposure adjust.} \\ 
		\hline
		\multirow{3}{*}{Trad. TMO}      & \cite{Marnerides18}  & \multirow{2}{*}{rand. param. TMOs} & \multicolumn{2}{c}{value $\sim U(0,15\%)$} \\ 
		\cline{2-2}\cline{4-5}
		& \cite{Cao21}         &                                    & $\times$   & $\times$     \\ 
		\cline{2-5}
		& \cite{Lee21,Borrego22}    & fixed param. TMOs             & $\times$   & $\times$     \\ 
		\hline
		Mid. exp. SDR                   & \cite{Jang18,Soh19,Jang20,Chambe22}    & fixed            & \checkmark & \checkmark  \\
		\hline
	\end{tabular}
\end{table}

First, nonlinearity between HDR and SDR is a monotonically increasing one-to-one mapping that itself will not introduce degradation. However, it substantially diverse between real-world cameras, and is measured as CRF\cite{CheatHDR}. Second, over/under-exposure truncation is to simulate the limited dynamic range of SDR. As in Table \ref{tab1}, there are 3 common practices to conduct above degradation:

`Simulated shot' means getting input SDR by applying virtual camera on the luminance recorded in label HDR. `Trad. TMO' is for traditional tone mapping operator (TMO) converting label HDR into input SDR. Finally, `Mid. exp. SDR' starts from a multi-exposure SDR sequence, where middle-exposure (EV=0) SDR is taken as input, meanwhile the whole sequence is merged as label HDR.

We argue that not all of them are favorable to SI-HDR training. For example, the motive of TMO is the exact opposite of degradation in that TMO dedicate to preserver as much information from HDR. In this case, DNN is not likely to learn adequate restoration ability since input SDR is also of high quality (see later experiment). Such analysis serves as the guidance of our training strategy.

%===========================================================
\section{Proposed Method}

%------------------------------------------------------------------------- 
\subsection{Problem Modeling}

As mentioned above, problem modeling is to figure out what degradations are to be restored, so we can (1) apply DNN mechanism tailored for specific degradation, and (2) arrange degradations correctly when training. Similar to \cite{Liu20}, we model the SI-HDR task as the reverse of HDR-to-SDR camera pipeline. Since their preliminary model could not envelop the source of noise and compression in legacy content, and the potential color space discrepancy between SDR and HDR (e.g. some HDR data\cite{FairchildHDR,HdMHDR} is in camera RGB primaries rather than sRGB), we derive a more comprehensive model from a precise camera pipeline\cite{BrownPipeline}.

\begin{figure}[b]
	\centering
	\includegraphics[width=\linewidth]{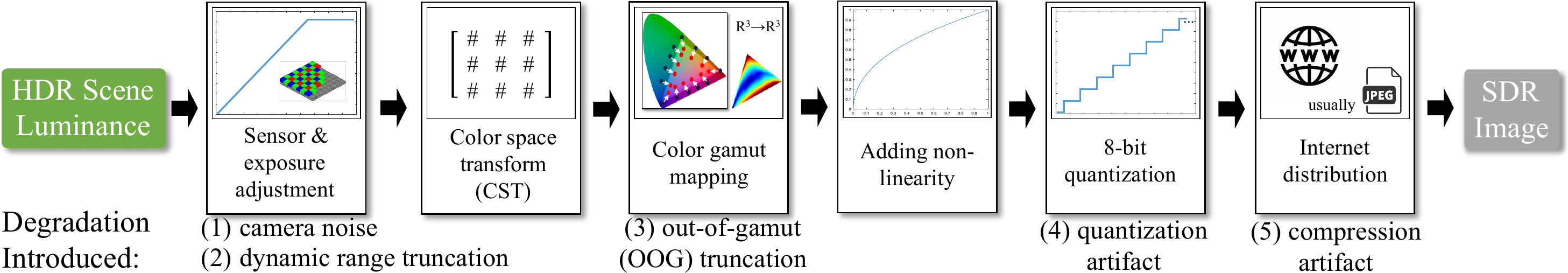} 
	\caption{Our problem modeling: SI-HDR is treated as the inverse of camera pipeline from linear HDR luminance to nonlinear SDR. See supplementary material for detailed derivation.}
	\label{fig1}
\end{figure}

As in Figure\ref{fig1}, our model consist of 6 steps with various degradations introduced sequentially. After determining 5 degradations and totally 7 operations to resolve in SI-HDR (nonlinearity and CST is not degradation, but do need an operation to reverse), one option is following \cite{Liu20} i.e. executing each step sequentially. However, cascaded sub-networks will make our method bulky thus conflicting with lightweight design (\cite{Liu20} is the slowest one, see Table\ref{tab3}). Hence, inspired by the taxonomy of traditional SI-HDR/EO(expansion operator), we turn to analyze each operation and divide them into 2 categories:

\textbf{Global operations.} CST and nonlinearity belong to global operation where neighboring pixels are not involved in its reversion. Also, it's proven in \cite{Liu21Retouching} that both CST and nonlinearity can be approcahed by a multi-layer perceptron (MLP) on multiple channels of single pixel.

\textbf{Local operations.} The rest are `local' operations where different extent of neighboring context is required in their reversion: From other low-level vision tasks e.g. denosing and compression artifact removal etc. we know that recovering (1)(3)(4)(5) in Figure\ref{fig1} only require the help of adjacent (small-scale) information. While under\&over-exposure reconstruction is more challenging since it requires long-distance dependency, similar as image inpainting.

%------------------------------------------------------------------------- 
\subsection{DNN Structure}

Since there are 2 distinct types of operation/degradation to resolve, we assign 2 steps of sub-network with different customized structure.

\begin{figure}[!b]
	\centering
	\subfloat[structure of global network]{\includegraphics[width=0.4\linewidth]{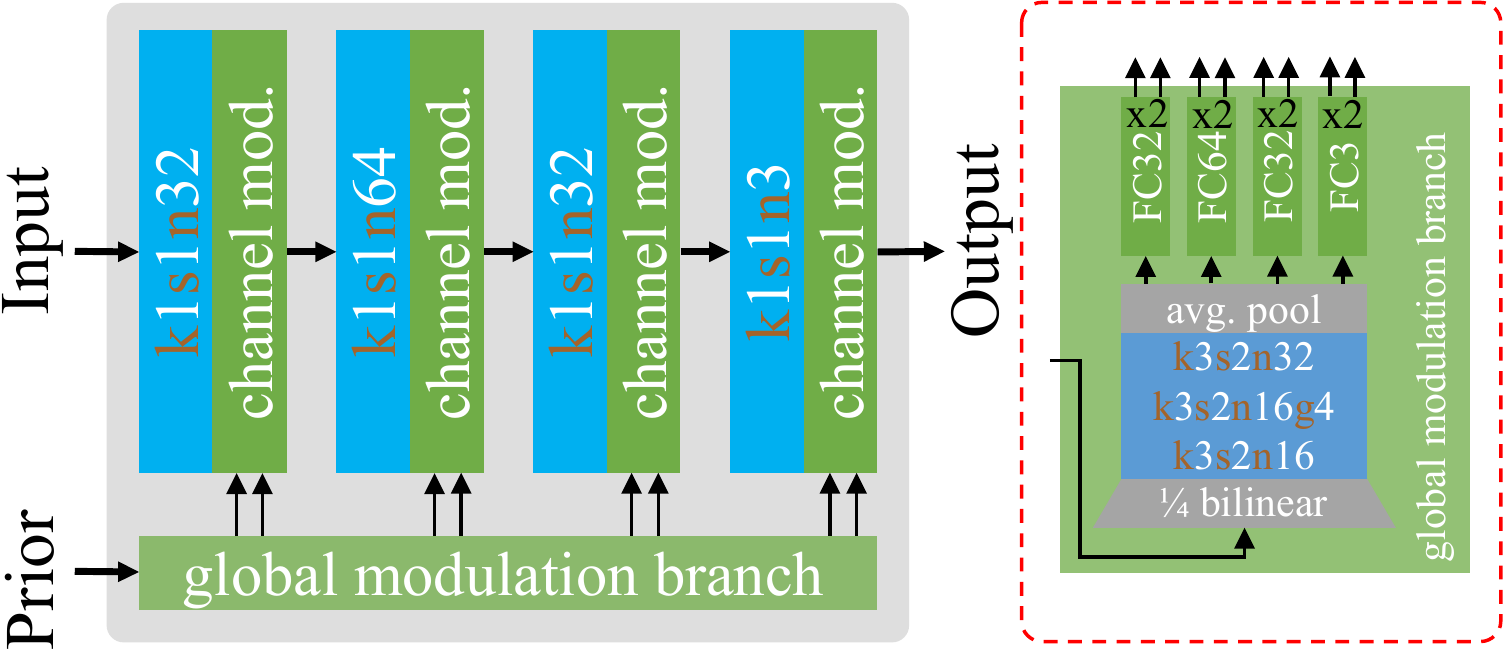}}
	\hfill
	\subfloat[structure of local network]{\includegraphics[width=0.55\linewidth]{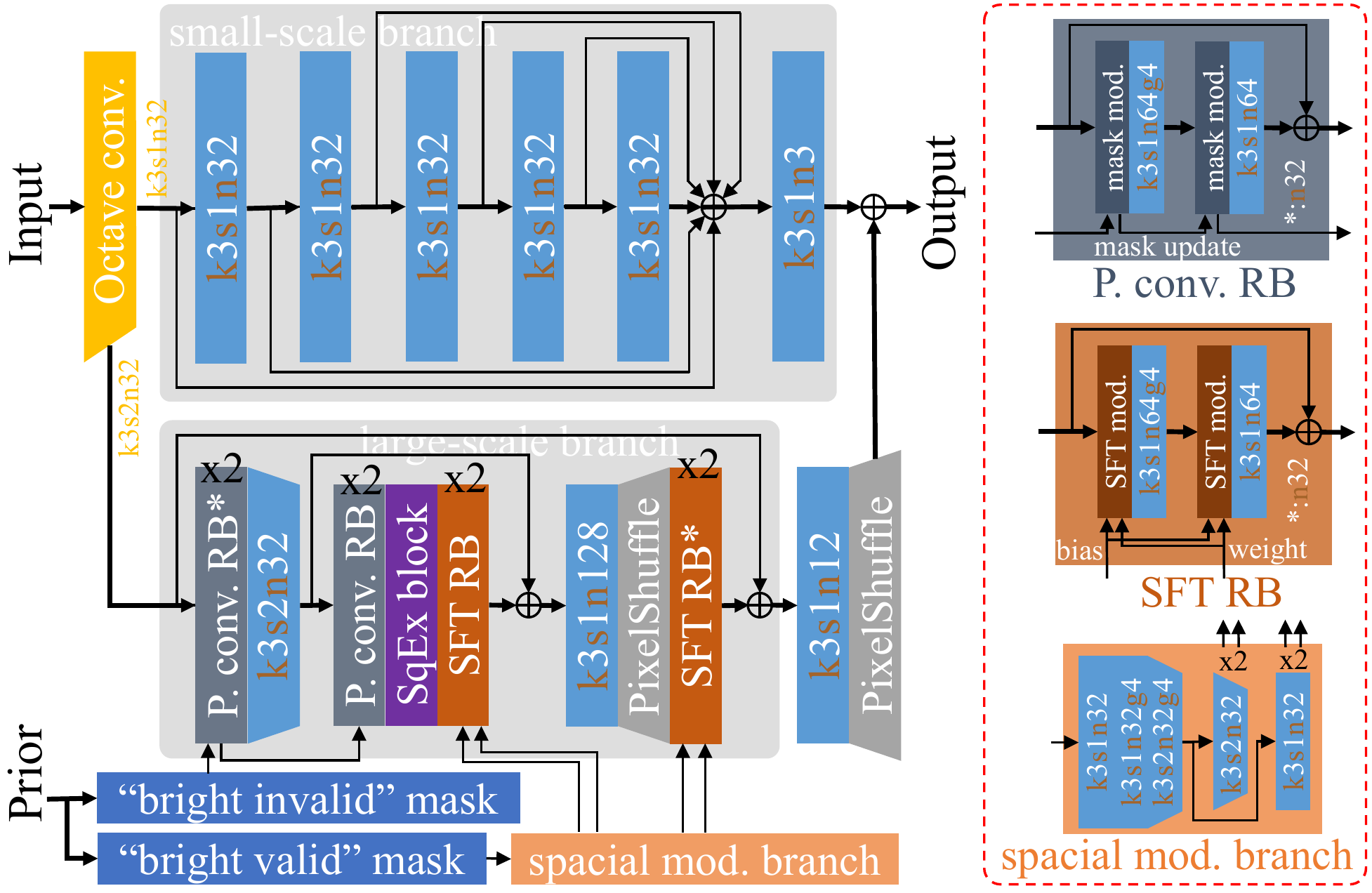}} \\
	\subfloat[final 2-step network]{\includegraphics[width=0.3\linewidth]{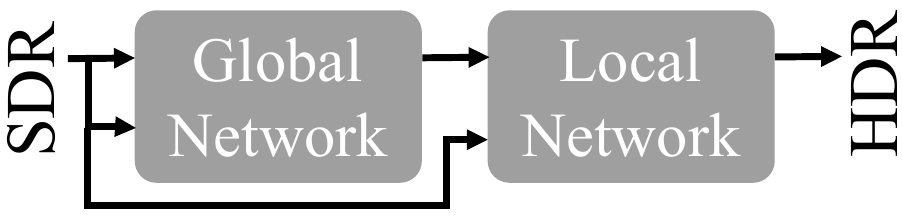}}
	\caption{DNN strcture of proposed method.}
	\label{fig2}
\end{figure}

\textbf{Global network}. While most DNN structures are capable of global operation, we adopt minimum-overhead MLP adhering to lightweight principle. Here, image per-pixel MLP is implemented by 4-layer point-wise ($1\times 1$) convolution.

In bottom Figure\ref{fig2}(a), prior with spatial-overall inforamtion is used to remedy insufficient receptive-filed.
Here, different from \cite{Raipurkar21}/\cite{Yu21} whose prior is segmentaion/attention map respectively, ours is input SDR itself depending on what prior information we need. Specifically, we need information of overall luminance/pixel energy. For example, an bright image should require less expansion in SI-HDR, and such prior will help the DNN to understand this. Finally, modulation branch will processe prior to $\mathbf{\alpha}_{scale},\mathbf{\beta}_{bias} \in \mathbb{R}^{c \times 1 \times 1}$ for modulation:
\begin{equation}
	modulation(\mathbf{x})=\mathbf{\alpha}_{scale} \odot \mathbf{x}+\mathbf{\beta}_{bias}, \mathbf{x} \in \mathbb{R}^{c \times h \times w}
	\label{eq1}
\end{equation}
where $\odot$ is channel-wise multiply. $\mathbf{\alpha}_{scale},\mathbf{\beta}_{bias} \in \mathbb{R}^{c \times 1 \times 1}$ ranther than $\mathbb{R}^{c \times h \times w}$ since it’s only supposed to contain spatial-global information. The superiority of prior's shape and type has been respectively proven in \cite{Chen21} and \cite{Liu21Retouching}.

\textbf{Local network} is responsible for inverting all local operations. We split it into 2 branches according to 2 scales of operation. In top Figure \ref{fig2}(b), we use a 5-layer densely-connected convolution block with small receptive-filed to deal with small-scale local operations. For large-scale under\&over-exposure reconstruction, we apply a 2-level encoder-decoder containing residual blocks (RB) to establish long-range dependency, at bottom Figure \ref{fig2}(b). We also utilize 2 types of soft-mask to ameliorate DNN's over-exposure recover capability: 

\begin{equation}
	\label{eq2}
	\left\{
	\begin{array}{lr}
		mask_{bright\ valid}(\mathbf{p})=max(0,(\mathbf{p}-t)/(1-t))\\
		mask_{bright\ invalid}(\mathbf{p})=max((\mathbf{p}-1)/(t-1), 1)
	\end{array}
    ,t=0.9, \mathbf{p}\in[0,1]
	\right.
\end{equation}

Here, prior ($\mathbf{p}$) cooperates with 2 kind of masks which aims to endow DNN with different spatial emphasis on over-exposed areas. Hence, the prior is also set to input SDR itself containing luminance information from where over-exposure can be inferred. Specifically, since the surrounding of saturated area is helpful for its recover (consider the case of center saturated direct illuminant), we put $mask_{bright\ valid}$ before spatial modulation (SFT\cite{SFT}) branch to reweight pixels there. SFT can also be expressed as Equation\ref{eq1}, but with $\mathbf{\alpha}_{scale},\mathbf{\beta}_{bias} \in \mathbb{R}^{c \times h \times w}$ and $\odot$ for pixel-wise multilpy. Yet, saturated area itself is of no useful information for its recover, we therefore multiply tensor with $mask_{bright\ invalid}$ before convolution (partial convolution\cite{PConv}), to exclude it from deep feature generation.

\textbf{Lightweight modules.} Apart from $1\times1$ convolution in global network with less \#param, we also change half of the convolution layers in local network's UNet branch to group($=4$) convolution. The main contributor of lightweight design lies, yet, in the idea that low-level vision task donot have an appetite for model complexity/depth. Finally, as in Figure \ref{fig2}(c), global and local network are integrated into a 2-step network. We put local network first since we empirically found it resulting more accurate color reproduction.Also, all activations are leaky rectified linear unit (leakyReLU) except the last layer (ReLU), hence we remove all normalization layers to fully utilize the nonlinear tail of leakyReLU.

%------------------------------------------------------------------------- 
\subsection{Training Strategy}
\label{sec:train}

We adopt supervised training where input SDR and traget HDR are required. As mentioned above, both HDR-exclusive and conventional degradations in input SDR images are crucial for the DNN's learned recover capability.

\begin{table}[!t]
	\centering
	\caption{Statistics of SDR images ($\mathbf{x}^\prime$) in candidate training set. Average portion (avg.) of under/over-exposure pixels ($i$) reflects the extent of HDR-exclusive degradation, its standard deviation (stdev.) stands for the degradation diversity. Note that over-exposure value (val.) of \cite{NTIRE21} is 248 i.e. $\nexists\ \mathbf{x}_i^\prime\in(248,255]$.}
	\label{tab2}
	\begin{tabular}{c|c|c|c|c|c|c|c|c|c} 
		\toprule
		\multirow{2}{*}{\textbf{Category}} & \multirow{2}{*}{\makecell[c]{{\textbf{Data}}\\{\textbf{set}}}} & \multirow{2}{*}{\makecell[c]{{\textbf{Resol}}\\{\textbf{ution}}}} & \multirow{2}{*}{\textbf{\#pair}} & \multicolumn{3}{l|}{\textbf{SDR under-exp. pixel}} & \multicolumn{3}{l}{\textbf{SDR over-exp. pixel}}    \\ 
		\cline{5-10}
		&                                                         &                                                                       &                          & \textbf{val.}   & \textbf{avg.(\%)} & \textbf{stdev.}   & \textbf{val.}                 & \textbf{avg.(\%)} & \textbf{stdev.}  \\ 
		\midrule
		\multirow{3}{*}{\begin{tabular}[c]{@{}c@{}}Simulated\\shot\end{tabular}} & \cite{Kalantari17MEHDR}         & 1.5k                                                                  & 74                       & \multirow{5}{*}{0} & 0         & N/A      & \multirow{2}{*}{255} & 3.533    & 0.0535  \\ 
		\cline{2-4}\cline{6-7}\cline{9-10}
		& \cite{Liu20}                                                   & 512                                                                   & 9786                     &                    & 6.365     & 0.1673   &                      & 8.520     & 0.1888  \\ 
		\cline{2-4}\cline{6-10}
		& \cite{NTIRE21}                                                 & HD                                                                    & 1494                     &                    & 8.840     & 0.0766   & 248                  & 4.711     & 0.0497  \\ 
		\cline{1-4}\cline{6-10}
		\multirow{2}{*}{\begin{tabular}[c]{@{}c@{}}Mid. exp.\\SDR\end{tabular}} & \cite{FairchildHDR}            & $>$4k                                                                  & 105                      &                    & 1.709     & 0.0647   & \multirow{2}{*}{255} & 4.355     & 0.0821  \\ 
		\cline{2-4}\cline{6-7}\cline{9-10}
		& \cite{Chambe22}                                                & 6k                                                                    & 400                      &                    & 1.200     & 0.0373   &                      & 1.081     & 0.0191  \\
		\hline
	\end{tabular}
\end{table}

\textbf{Dataset with HDR-exclusive degradation.} Many previous works have released their training set in the form of HDR-SDR pairs, which means HDR-exclusive degradations are already contained. Those datasets have crafted highly-diversified scenes, and are generated by 1 of the 3 ways in Table\ref{tab1}. As analyzed there, `Trad. TMO' is not beneficial, hence we exclude all such training set from candidate. In Table\ref{tab2}, we quantify the statistics of degraded SDR images in candidate training set. We finally mix NTIRE\cite{NTIRE21} and Fairchild\cite{FairchildHDR} Dataset based on such statistics: that their over\&under-exposure degradation is of appropriate extent and diversity. Meanwhile, the HDR-SDR nonlinearity in \cite{FairchildHDR} is fixed, this is remedied since that of \cite{NTIRE21} is diversified, as in Table\ref{tab1}. Their ratio in final training patches (sized $600\times600$, total 6060) is about 11:4(NTIRE:Fairchild).

\textbf{Conventional degradations} are to simulate the characteristics of legacy SDR. We conduct them on off-the-shelf SDR images from above datasets: From \cite{BrownPipeline} we know that camera noise is first gathered at sensor's linear RAW response, while compression is lastly added before storage. Hence, we first linearize SDR image and then simulate it to camera RAW RGB primaries, before adding Poisson-Gaussian camera noise whose distribution is simplified to heteroscedastic Gaussian with $\sigma \sim\ U(0.001,0.003)$. Then, the image is transferred back to sRGB, before adding compression. Like \cite{DoubleJPEG}, we use double JPEG compression to simulate multiple internet transmissions of legacy content, but we use a more realistic model where the first QF$\sim U(60,80)$ and the second QF is fixed 75 but on rescaled image patches. See supplementary material for detail.

\textbf{Data pre\&post-processing} can make pixel value of linear HDR more evenly-distributed thus easier for DNN to understand. While various non-linear pre-processing e.g. $\mu$-law etc. are widely exploited\cite{Eilertsen17,Santos20,Zhang17,Jang18,Wang19,Soh19,Marnerides21,Zhang21,Yu21}, we choose a simple-yet-effective gamma pre-processing: All label HDR images ($\mathbf{y}$) are normalized by their maximum recorded luminance, then transferred to nonlinear domain (denote with superscript $\prime$) before sending to DNN, i.e.

\begin{equation}
	\mathbf{y}^\prime=(\frac{\mathbf{y}}{max(\mathbf{y})})^{\gamma},\gamma=0.45
	\label{eq3}
\end{equation}
In this case, we append post-processing $\bar{\mathbf{y}}={{\bar{\mathbf{y}}}^\prime}{}^{1/0.45}$ on DNN's output HDR image (${\bar{\mathbf{y}}}^\prime$) during inference phase, to bring it back to linear domain.

\textbf{Loss function.} Our `$l_1$' and `$l_g$' (gradient loss) can both be formulated as average element($i$)-wise distance: $\frac{1}{n} \sum_{i=1}^{n}{\left|| f(\bar{\mathbf{y}}_i^\prime)-f(\mathbf{y}_i^\prime) \right||}_1$ where $f()$ is non-op for `$l_1$', and a discrete differential operator (outputting $\mathbb{R}^{2\times 3\times h\times w}$ where 2 means gradient map on both horizontal and vertical direction) for `$l_g$'. The latter is added to highlight the local structure perturbation brought by noise and compression artifact. Final loss are empirically set to: $l_1+0.1\times l_g$. All DNN parameters are Kaiming initialized and optimized by AdAM with learning rate starting from $2\times10^{-4}$ and decaying to half every $2.5\times10^5$ iters. More training details can be found in supplementary material.

%===========================================================
\section{Experiments}

\textbf{Criteria.} Accroding to the motive of our SI-HDR method, we focus on 3 aspects: (1) Recovery ability of lost information in saturated area, (2) lumiance estimation accuracy and (3) restoration capability of conventional degradations. The first two apply to all SI-HDR methods, while the last one is our emphasis.

Moreover, as reported by \cite{CheatHDR} and \cite{Hanji22}, current pixel-distance-based metrics tend to make an assessment paradoxical to subjective result. Hence, we introduce more detailed portrait to assess (1) and (3), while metrics are still used for (2).

%------------------------------------------------------------------------- 
\subsection{Comparison with SOTA}

Our method is compared with 8 others including 7 DNN-based and 3 with lightweight motive, as listed in Table\ref{tab3}. We use PSNR, SSIM and `quality correlate' from HDR visual difference predictor 3 (VDP)\cite{VDP3} for quantitative analysis.

\textbf{On simulated legacy SDR.} The test set consists of 95 simulated legacy SDR images from same dataset and extent of degradation as in training. Note that some methods will inevitably struggle to handle SDR with noise and compression, since they are not trained so. Yet, we didn't re-train them because it's not our work to compensate for their insufficiency in training strategy e.g. \cite{Marnerides18} has weaker over-exposure recovery ability mainly due to HDR-exclusive degradation in training set. Still, for a fair comparison, for methods not trained to handle such conventional degradations, we first process input SDR with pre-trained auxiliary DNN to mitigate compression artifact($\star$)\cite{DnCNN-JPEG} and/or noise($\dagger$)\cite{DnCNN}.

\begin{table}[!t]
	\centering
	\caption{Quantitative comparison of all competitors. Note that MACs and runtime (both for $\mathbb{R}^{3 \times 1080 \times 1920}$ input) are counted only within DNN, and on desktop computer with i7-4790k CPU and GTX1080 GPU, respectively. Superscript$^{1/2/3}$ stands for Tensorflow/PyTorch/MATLAB implementation, `DeC.' and `DeN.' is respectively for compression artifact removal and denoise.}
	\label{tab3}
	\begin{tabular}{c|c|c|c|r|r|r|r|r|r} 
		\toprule
		\multirow{2}{*}{\textbf{Method}} & \multicolumn{2}{c|}{\textbf{Designed to}}  & \multirow{2}{*}{\makecell[c]{\textbf{Lightw.} \\ \textbf{Motive}}} & \multicolumn{3}{c|}{\textbf{Overhead}} & \multicolumn{3}{c}{\textbf{Metrics}}  \\ 
		\cline{2-3}\cline{5-10}
		& DeC.                 & DeN.                &                            & \#param & MACs  & runtime                & PSNR  & SSIM   & VDP       \\ 
		\midrule
		HDRCNN\cite{Eilertsen17} & \checkmark          & $\times^{\dagger}$  & $\times$                   & 31799k  & 2135G & $15.54s^1$                & 32.40 & .8181 & 5.806       \\ 
		\hline
		ExpandNet\cite{Marnerides18} & $\times^{\star}$ & $\times^{\dagger}$ & $\times$                   & 457k    & 508G  & $0.88s^2$                  & 20.47 & .5915 & 4.653       \\ 
		\hline
		SingleHDR\cite{Liu20}    & $\times^{\star}$    & $\times^{\dagger}$  & $\times$                   & 30338k  & 1994G & $41.98s^1$ & \underline{35.52} & \underline{.9100} & \underline{6.231} \\ 
		\hline
		DHDR\cite{Santos20}      & $\times^{\star}$    & $\times^{\dagger}$  & $\times$                   & 51542k  & 597G  & $10.49s^2$                 & 20.00 & .4760 & 4.766       \\ 
		\hline
		HDRUNet\cite{Chen21}     & $\times^{\star}$    & \checkmark          & $\times$                   & 1651k   & 741G  & $1.51s^2$                  & 31.06 & .7638 & 5.688       \\ 
		\hline
		FHDR\cite{Khan19} & $\times^{\star}$    & $\times^{\dagger}$  & \checkmark                 & 571k    & 2368G & $10.60s^2$                 & 20.26 & .4374 & 4.985       \\ 
		\hline
		HDR-LFNet\cite{Chambe22} & $\times^{\star}$    & $\times^{\dagger}$  & \checkmark    & \textbf{203}k & \textbf{47.7}G & $2.94s^2$                & 25.73 & .4536 & 4.791       \\ 
		\hline
		RempelEO\cite{Rempel07EO} & \checkmark         & \checkmark          & \checkmark     & \multicolumn{2}{c|}{N/A}    & $6.75s^3$                   & 24.14 & .8045 & 4.710       \\ 
		\hline
		ours                     & \checkmark          & \checkmark        &  \checkmark  & \underline{225}k & \underline{162}G  & \textbf{0.53}$s^2$ & \textbf{38.12} & \textbf{.9515} & \textbf{7.173} \\ 
		\hline
	\end{tabular}
\end{table}

\begin{figure}
	\centering
	\subfloat[Detailed comparison. All HDR images/patches in relative linear luminance are visualized by MATLAB tone-mapping operator (TMO) `localtonemap'. We select this TMO since it preserves local contrast thus detail information in both dark and bright areas become more conspicuous.]{\includegraphics[width=\linewidth]{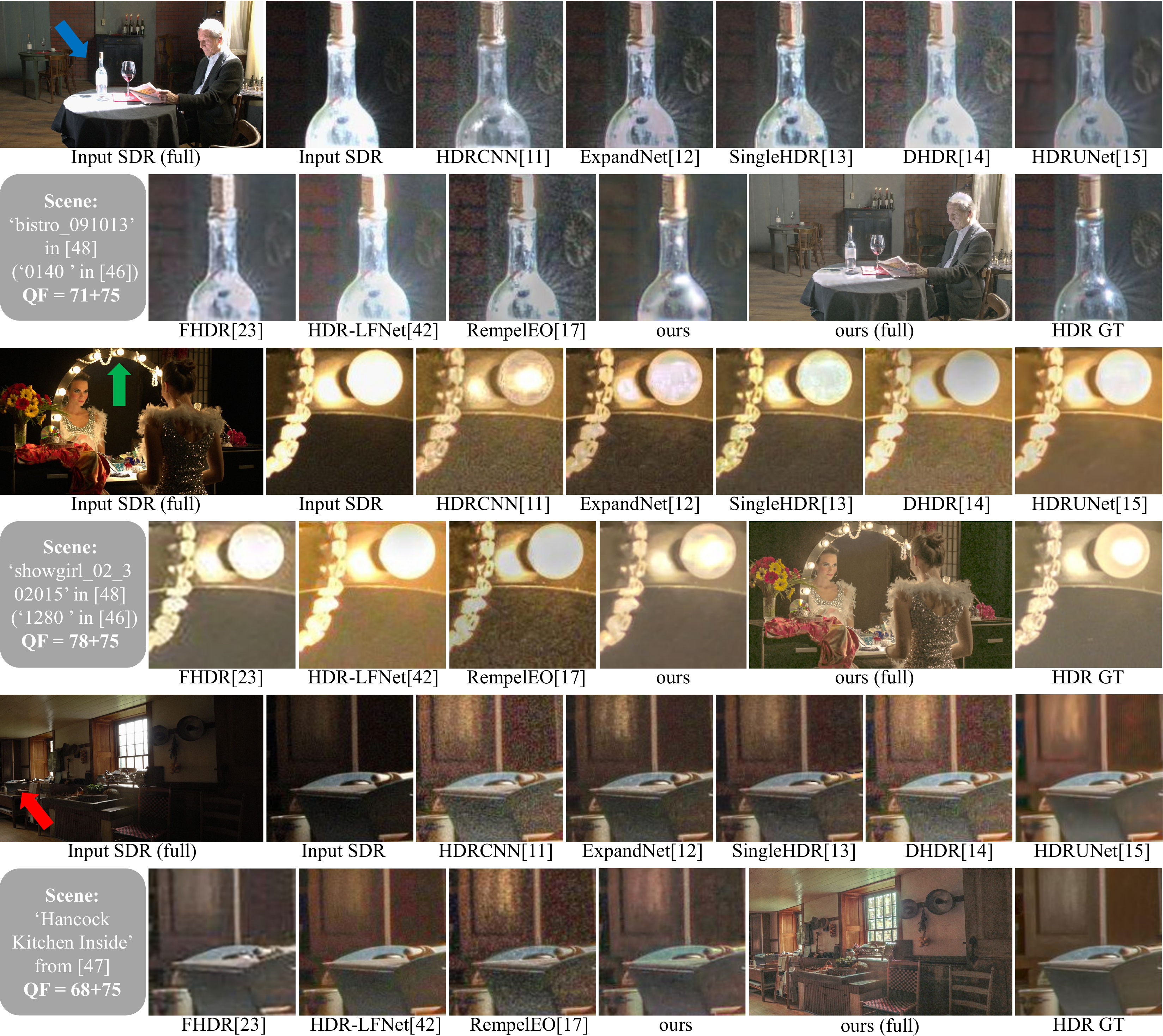}} \\
	\subfloat[Recovered HDR luminance, visualized by MATLAB heatmap `turbo'. Closer luminance distribution with GT means better for IBL\cite{ReinhardHDRBook} application.]{\includegraphics[width=\linewidth]{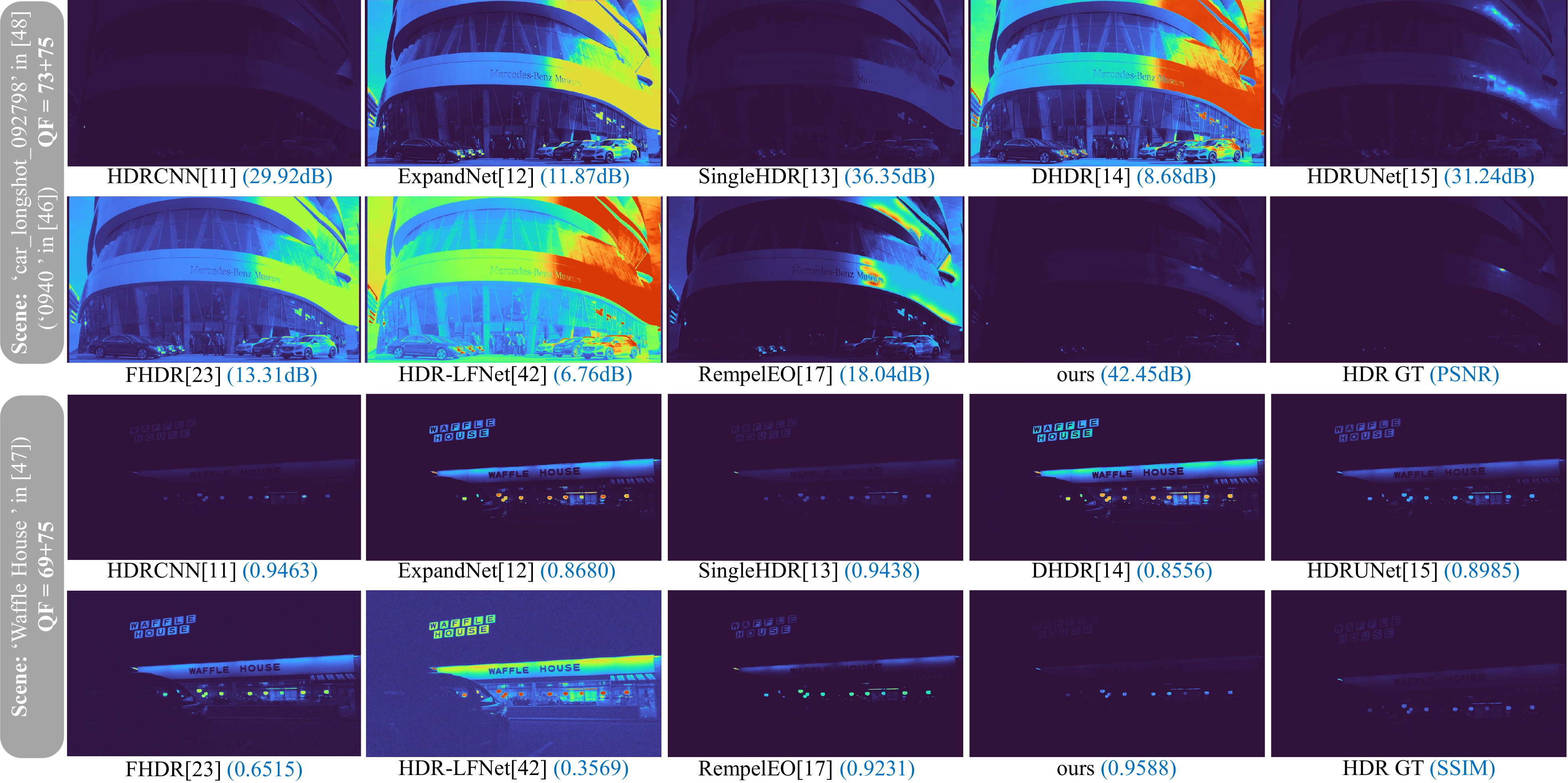}}
	\caption{Results on simulated legacy SDR input.}
	\label{fig4}
\end{figure}

In Table\ref{tab3}, our method got the highest metrics with minimal runtime and second least \#param and MACs. As reported by \cite{CheatHDR}, the main contributor to an appealing score lies in method's estimation accuracy of nonlinearity/CRF. This idea can be confirmed by the heatmap in Figure\ref{fig4}(b) where each method's luminance distribution corespond with metrics. Therefore, we turn to assess degradation recovery ability by detailed visual comparison in Figure\ref{fig4}(a). As seen, our method is able to suppress noise and compression while recover adequate information from dark\&saturated areas.

\begin{figure}[!t]
	\centering
	\includegraphics[width=\linewidth]{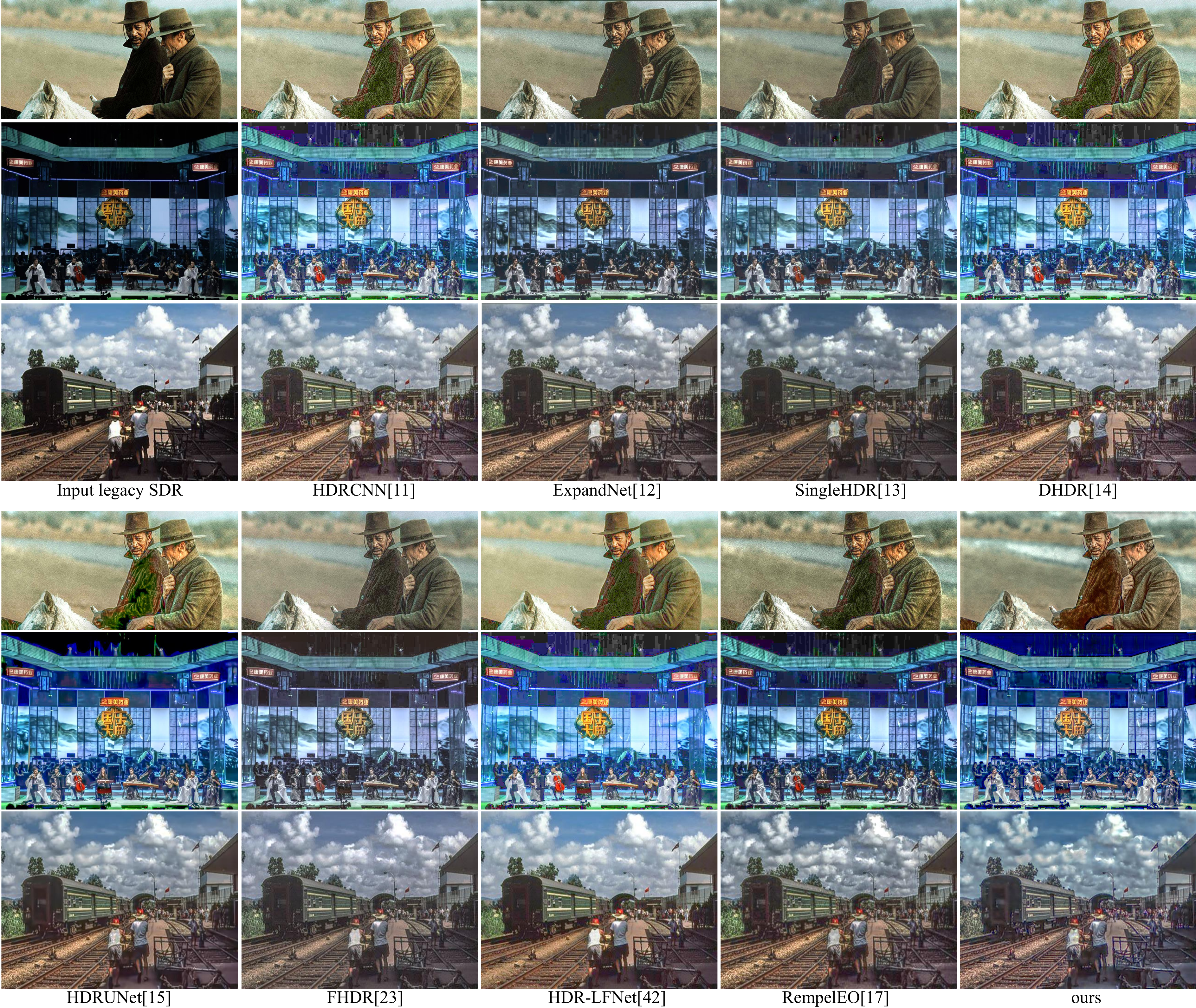} 
	\caption{Results on real legacy SDR input, tone-mapped for visualize.}
	\label{fig5}
\end{figure}

\textbf{On real legacy SDR.} Legacy SDR in real application scenarios usually contains an unknown degree of (blind) degradation. Here, each method is tested on 51 real legacy SDR images from old movies, photographs, and TV programs. Since there is no GT counterpart, we only provide visual comparison in Figure\ref{fig5}. As seen, our method is able to mitigate noise and compression while avoiding producing artifacts at dark area which is more susceptible to noise.

\textbf{Analysis.} First, albeit some methods hold lightweight motive, they still take long a time to run because: only \#param is reduced while MACs is not\cite{Khan19}; most runtime is spent in data pre/post-procesing outside the DNN\cite{Chambe22}.

Meanwhile, from Figure\ref{fig4} we can see that most competitors fail to jointly denoise and remove compression artifact even with the help of auxiliary DNN ($\star/\dagger$), and result with less artifact only appears in \cite{Eilertsen17}/\cite{Chen21} which is also respectively trained with corresponding degradation. This confirms the significance conventional degradation in training set.

Also, some methods \cite{Marnerides18,Chambe22} underperform reconstructing over-exposure area. Reanson for \cite{Chambe22} is that their DNN just polishes the result of traditional expansion operators (EOs), from where the lost information was never recovered. For \cite{Marnerides18}, the only competitor who is trained with `Trad. TMO' dataset (see Table\ref{tab1}), reason lies in the insufficient degradation ability of ‘Trad. TMO’ ‘degradation’. This proves the importance of HDR-exclusive degradation model.

The improtance of degradation model etc. will be further proved in ablation studies below. More heuristics analysis, visual results, and detailed experiment configuration are also provided in supplementary material.

%------------------------------------------------------------------------- 
\subsection{Ablation Studies}

In this section, we strat to verify the effectiveness of some key ingredients of our method, including training configuration and DNN structure.

\begin{figure}[!b]
	\centering
	\subfloat[The impact of HDR-exclusive degradation on over-({\color{green} green arrow}) and under-exposure({\color{red} red arrow}) recovery ability.]{\includegraphics[width=0.49\linewidth]{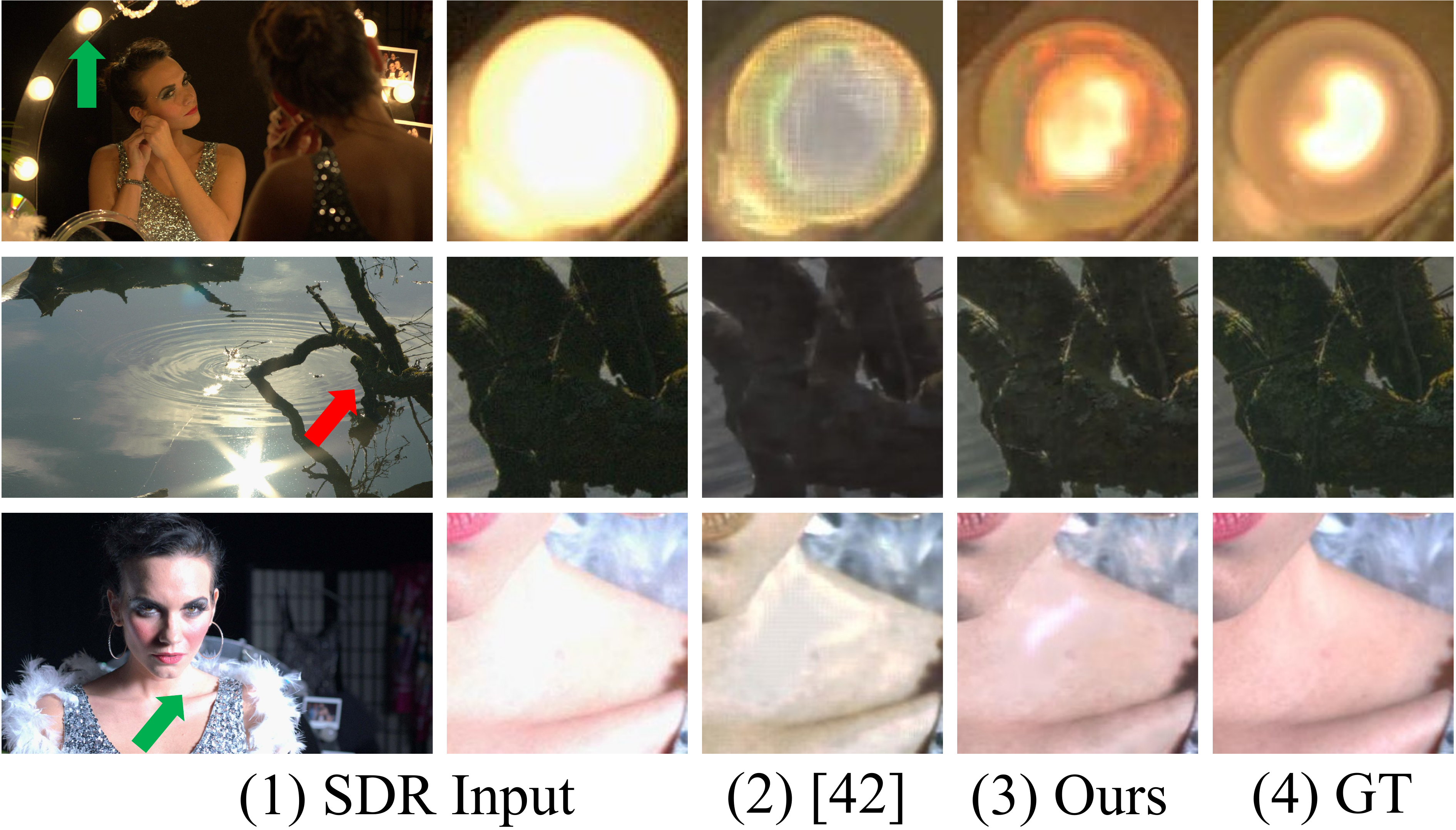}}
	\hfill
	\subfloat[The effect of conventional degradations on artifact restoration capability.]{\includegraphics[width=0.49\linewidth]{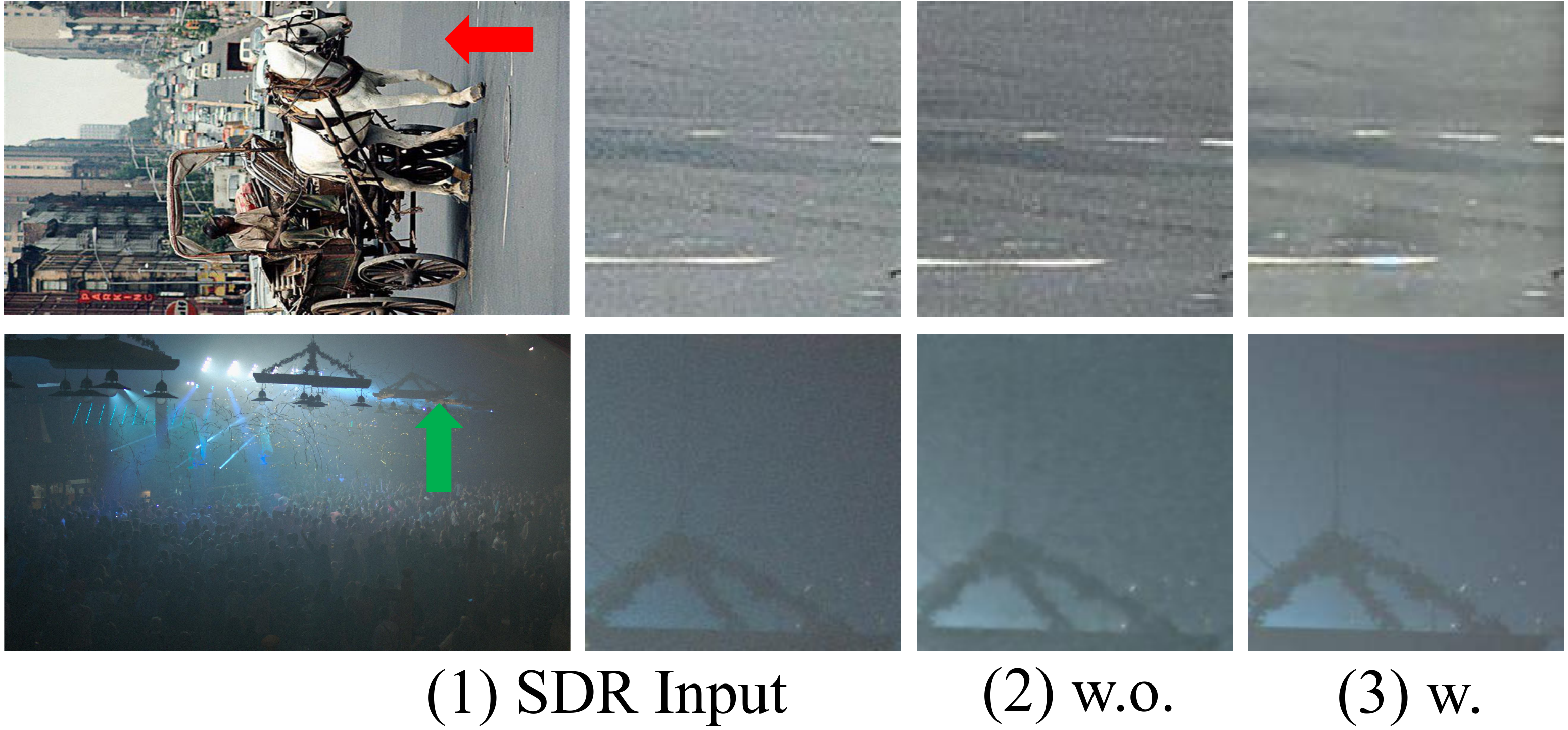}}
	\\
	\subfloat[on partial convolution\cite{PConv}]{\includegraphics[width=0.49\linewidth]{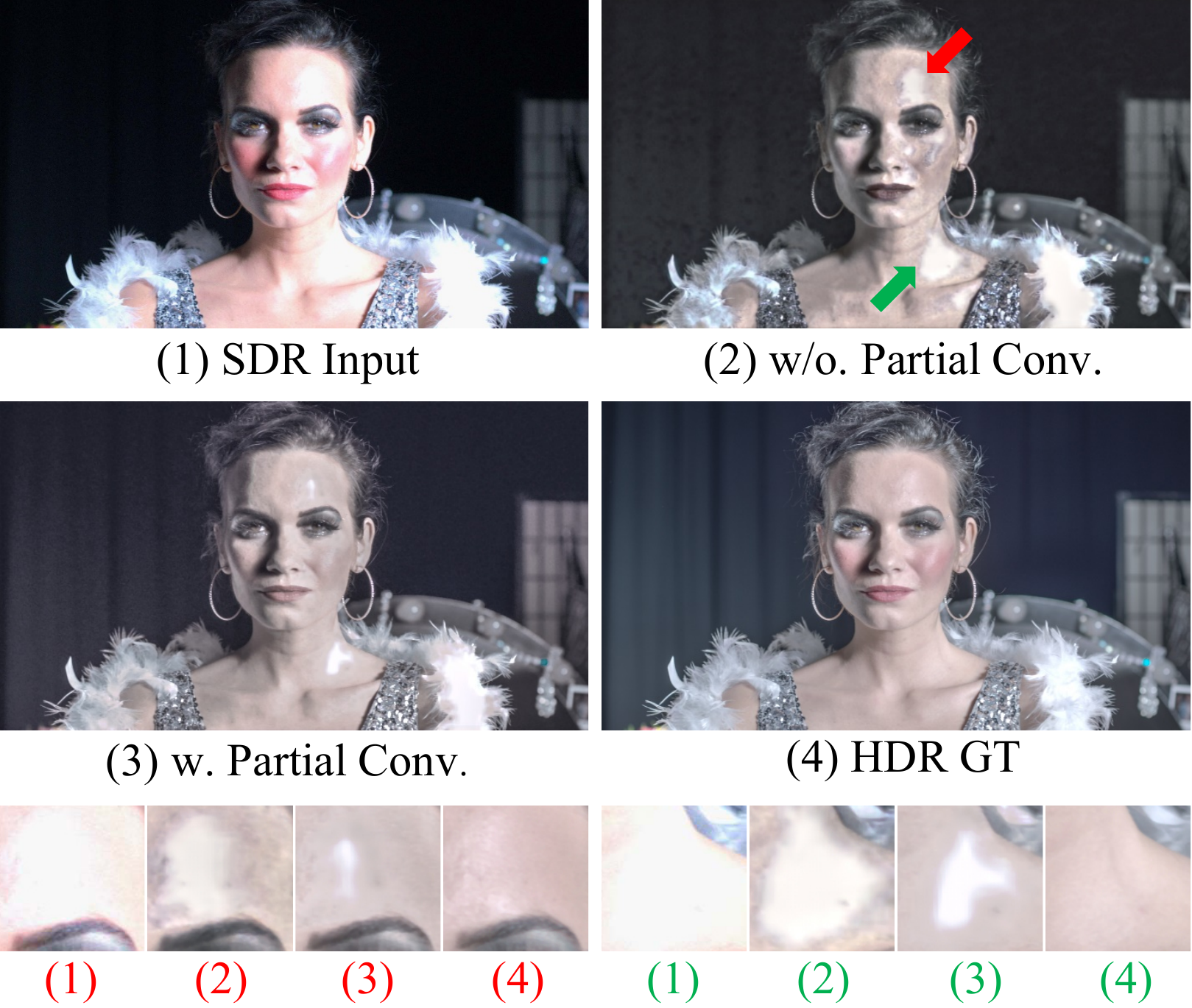}}
	\hfill
	\subfloat[on group convolution]{\includegraphics[width=0.49\linewidth]{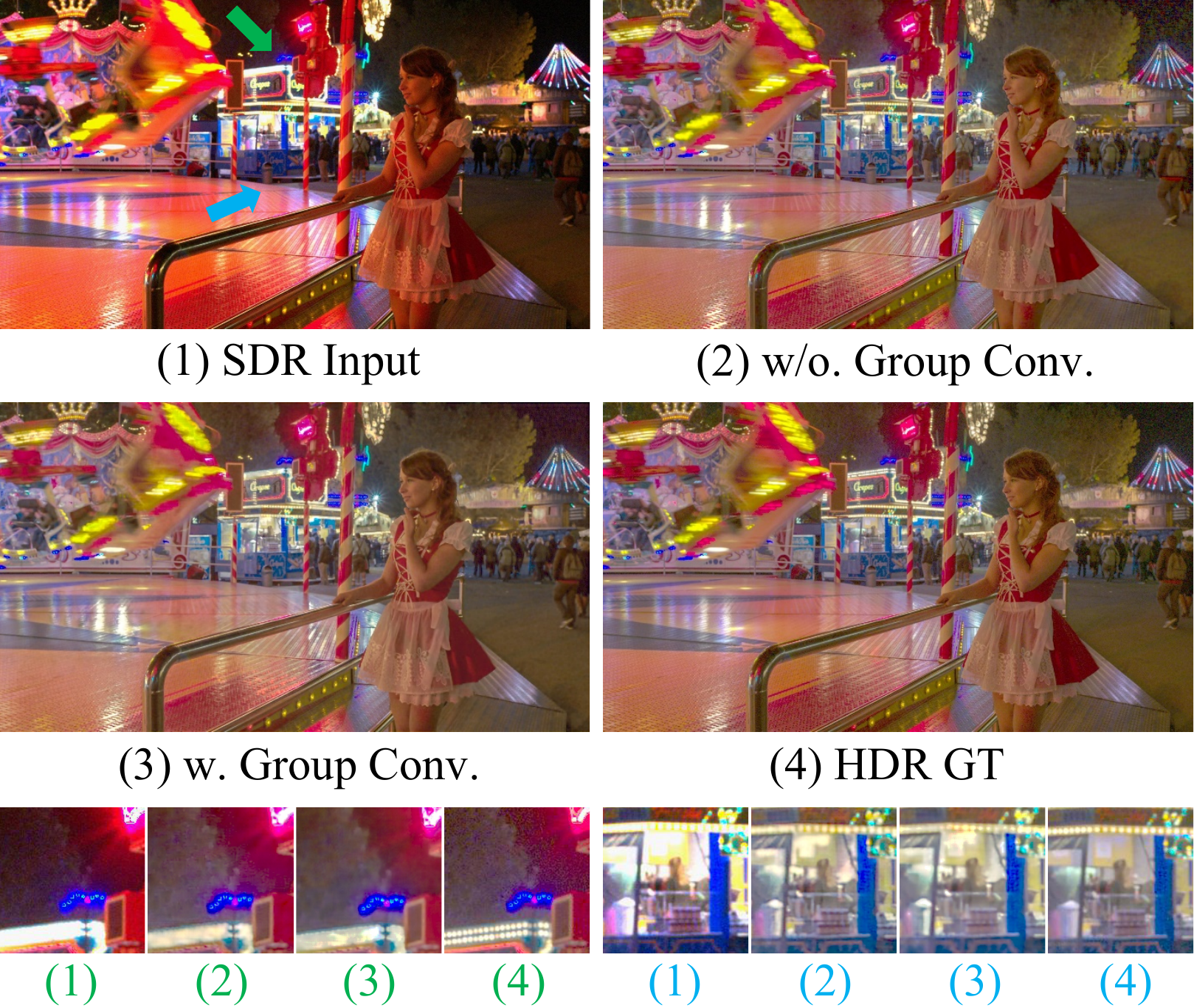}}
	\caption{Visual demonstration of ablation studies.}
	\label{fig3}
\end{figure}

\begin{table}[!b]
	\centering
	\caption{Metrics of ablation studies. `Con. deg.' stands for conventional degradations i.e. camera noise and JPEG compression as in Section\ref{sec:train}.}
	\label{tab4}
	\begin{tabular}{c|c|c|c|r|r|r|r|r|r} 
		\toprule
		\multicolumn{4}{c|}{\textbf{Ablation Configuration}}  & \multicolumn{3}{c|}{\textbf{Overhead}} & \multicolumn{3}{c}{\textbf{Metrics}}  \\ 
		\hline
		No. & DNN struct.         & Training set           &  Con. deg.     & \#param & MACs  & runtime    & PSNR  & SSIM  & VDP       \\ 
		\midrule
		\multicolumn{4}{c|}{unchanged (denote with -)}  & \multirow{3}{*}{225k}  & \multirow{3}{*}{162G} & \multirow{3}{*}{0.53s}      & 38.12 & .9515 & 7.173       \\
		\cline{1-4} \cline{8-10} 
		\textcircled{1} & -                   & change to \cite{Chambe22}    & -           &         &       &          & 29.55 & .7123 & 5.370      \\ 
		\cline{1-4} \cline{8-10} 
		\textcircled{2} & -                   & -                      & w/o.              &         &       &          & 37.16 & .8403 & 6.786       \\ 
		\hline
		\textcircled{3} & w/o. P.C.           & -                      & -                 & 337k    & 206G  & 0.66s      & 36.29 & .8112 & 6.603    \\ 
		\hline
		\textcircled{4} & w/o. G.C.           & -                      & -                 & 555k    & 207G  & 0.69s      & 38.85 & .9570 & 7.191     \\ 
		\hline
	\end{tabular}
\end{table}

\textcircled{1}\textbf{On HDR-exclusive degradation.} In Section\ref{sec:hdrdm}\&\ref{sec:train}, we argue that a lower degree of over/under-exposure degradation of training SDR tend to endow DNN with less recovery ability. From Table\ref{tab2} we know that the proportion of over/under-exposure pixels in current training set \cite{NTIRE21,FairchildHDR} is about 2-9\%. Here, keeping other variables unchanged, we replace the training set with HDR-LFNet\cite{Chambe22} whose SDR over/under-exposure pixels only account about 1\% of the image (see Table\ref{tab2}). As anticipated, when trained with sparingly-degraded SDR,  Figure\ref{fig3}(a)(2) recover far less content in over\&under-exposed areas than (a)(3). Meanwhile, metrics in Table\ref{tab4}\textcircled{1} drop significantly since DNN learned a different lumiance value distribution from another training set.

\textcircled{2}\textbf{On conventional degradations.} Here, we remove all camera noise and JPEG compression, to see if DNN still learn corresponding restoration ability. As in Figure\ref{fig3}(b)(2), without extra degradations in training set, our method also struggle to suppress noise and compression artifact, same as other methods without conventional training degradtions. Also, in Table\ref{tab4}\textcircled{2}, PSNR drop slightly since the training set is unchanged, thus the learned pixel energy distribution is still accurate. While the decline of SSIM and VDP-Q is relatively larger since the local structure is perturbed by noise and compression artifact.

\textcircled{3}\textbf{On partial convolution and bright invalid mask.} They are claimed as another contributor to DNN's over-exposure recover ability, apart from HDR-exclusive degradation above. Here, we show their indispensability by replacing it with a symmetric structure of decoder, i.e. `P.conv. RB' at bottom Figure\ref{fig2}(b) are replaced with `SFT RB'. In this case, to-be-recover saturated pixels in Figure\ref{fig3}(c) tend to spread, and metrics also suffer a decline as `w/o. PC' in Table\ref{tab3}\textcircled{3} shows. The immediate effect proves their ability to exclude useless saturated pattern hence better intermediate deep feature will be generated for the decoder's reconstruction. This is the exact reason why they are placed at encoder rather than decoder end.

\textcircled{4}\textbf{On group convolution.} Lightweight DNN has been proven of adequate capability for other low-level vision tasks\cite{NTIRE22MEHDR,AIM20ESR,NTIRE22ESR,MobileAI21FDN}. Therefore, we want to check if this also makes sense in SI-HDR. Group convolution (G.C.) is one of the contributor of lightweight design, here we depict if it will deteriorate the DNN's perfromance. By comparing Table\ref{tab3}\textcircled{4} we know that lightening DNN using group convolution do cause a slight decline on metrics, however, will lead to few noticeable difference in visual result, as Figure\ref{fig3}(d) shows.

%===========================================================
\section{Conclusion}

Legacy SDR content contains classic historical scenes that cannot be reproduced, and, however, limited dynamic range and degradations brought by old imaging system and multiple transmissions. This formulates an ill-posed problem, SI-HDR, when we want to put those legacy content into HDR application.

The community has begun to take advantage of DNN in SI-HDR problem. We also used DNN but handled more specific problems that hinder current DNN-based methods from real-world deployment: we designed a more lightweight DNN and trained it with elaborately designed degradations. Meanwhile, we reformed SI-HDR problem modeling to better derive DNN structure and arrange degradations correctly. Experiments show that our method is readily applicable to both synthetic and real legacy SDR content. Ablation studies also reveal some factor that will significantly impact the performance of SI-HDR method, including degradation model which has long been understated. Our code is available\footnote{\url{https://www.github.com/AndreGuo/LHDR}}.

Despite the preliminary step we made towards legacy-content-applicable SI-HDR, its cross-degradation generalizability still call for improvement: First, our method perfrom well on degraded legacy content, but not on clean SDR. Specifically, it tend to vanish/over-smooth high-frequency detail which is mistaken as degraded pattern. This issue also occurs in \cite{Chen21} etc. and should be considered by all legacy-SDR-oriented SI-HDR methods.

\subsubsection{Acknowledgements} This work was supported by the PCL2021A10-1 Project of Peng Cheng Laboratory.

% \bibliography{egbib}

\begin{thebibliography}{8}
	\bibitem{ReinhardHDRBook}
	Reinhard, E., Heidrich, W., Debevec, P.,  et~al.:
	\newblock High dynamic range imaging: acquisition, display, and image-based
	lighting.
	\newblock Morgan Kaufmann (2010)
	
	\bibitem{Debevec97}
	Debevec, P.E., Malik, J.:
	\newblock Recovering high dynamic range radiance maps from photographs.
	\newblock In: Proc. SIGGRAPH '97. (1997)  369–378
	
	\bibitem{Kalantari17MEHDR}
	Kalantari, N.K., Ramamoorthi, R.,  et~al.:
	\newblock Deep high dynamic range imaging of dynamic scenes.
	\newblock ACM Trans. Graph. \textbf{36} (2017)  144--1
	
	\bibitem{Wu18MEHDR}
	Wu, S., Xu, J.,  et~al.:
	\newblock Deep high dynamic range imaging with large foreground motions.
	\newblock In: Proc. ECCV. (2018)  117--132
	
	\bibitem{Yan19MEHDR}
	Yan, Q., Gong, D., Shi, Q.,  et~al.:
	\newblock Attention-guided network for ghost-free high dynamic range imaging.
	\newblock In: Proc. CVPR. (2019)  1751--1760
	
	\bibitem{Yan20MEHDR}
	Yan, Q., Zhang, L., Liu, Y.,  et~al.:
	\newblock Deep hdr imaging via a non-local network.
	\newblock IEEE Trans. Image Process. \textbf{29} (2020)  4308--4322
	
	\bibitem{Chen21MEHDR}
	Chen, G., Chen, C., Guo, S.,  et~al.:
	\newblock Hdr video reconstruction: A coarse-to-fine network and a real-world
	benchmark dataset.
	\newblock In: Proc CVPR. (2021)  2502--2511
	
	\bibitem{NTIRE22MEHDR}
	P{\'e}rez-Pellitero, E.,  et~al.:
	\newblock Ntire 2022 challenge on high dynamic range imaging: Methods and
	results.
	\newblock In: Proc. CVPR. (2022)  1009--1023
	
	\bibitem{Banterle06ITM}
	Banterle, F., Ledda, P., Debattista, K.,  et~al.:
	\newblock Inverse tone mapping.
	\newblock In: Proceedings of the 4th international conference on Computer
	graphics and interactive techniques in Australasia and Southeast Asia. (2006)
	349--356
	
	\bibitem{CheatHDR}
	Eilertsen, G., Hajisharif, S.,  et~al.:
	\newblock How to cheat with metrics in single-image hdr reconstruction.
	\newblock In: Proc. ICCV. (2021)  3998--4007
	
	\bibitem{Eilertsen17}
	Eilertsen, G., Kronander, J.,  et~al.:
	\newblock Hdr image reconstruction from a single exposure using deep cnns.
	\newblock ACM Trans. Graph. \textbf{36} (2017)  1--15
	
	\bibitem{Marnerides18}
	Marnerides, D., Bashford-Rogers, T.,  et~al.:
	\newblock Expandnet: A deep convolutional neural network for high dynamic range
	expansion from low dynamic range content.
	\newblock Comput. Graph. Forum \textbf{37} (2018)  37--49
	
	\bibitem{Liu20}
	Liu, Y.L., Lai, W.S.,  et~al.:
	\newblock Single-image hdr reconstruction by learning to reverse the camera
	pipeline.
	\newblock In: Proc. CVPR. (2020)  1651--1660
	
	\bibitem{Santos20}
	Santos, M.S., Ren, T.I., Kalantari, N.K.:
	\newblock Single image hdr reconstruction using a cnn with masked features and
	perceptual loss.
	\newblock ACM Trans. Graph. \textbf{39} (2020)  80--1
	
	\bibitem{Chen21}
	Chen, X., Liu, Y.,  et~al.:
	\newblock Hdrunet: Single image hdr reconstruction with denoising and
	dequantization.
	\newblock In: Proc. CVPR. (2021)  354--363
	
	\bibitem{BrownPipeline}
	Karaimer, H.C., Brown, M.S.:
	\newblock A software platform for manipulating the camera imaging pipeline.
	\newblock In: Proc. ECCV. (2016)  429--444
	
	\bibitem{Rempel07EO}
	Rempel, A.G., Trentacoste, M.,  et~al.:
	\newblock Ldr2hdr: on-the-fly reverse tone mapping of legacy video and
	photographs.
	\newblock ACM Trans. Graph. \textbf{26} (2007)  39--es
	
	\bibitem{Zhang17}
	Zhang, J., Lalonde, J.F.:
	\newblock Learning high dynamic range from outdoor panoramas.
	\newblock In: Proc. ICCV. (2017)  4519--4528
	
	\bibitem{Jang18}
	Jang, H.,  et~al.:
	\newblock Inverse tone mapping operator using sequential deep neural networks
	based on the human visual system.
	\newblock IEEE Access \textbf{6} (2018)  52058--52072
	
	\bibitem{Moriwaki18}
	Moriwaki, K., Yoshihashi, R.,  et~al.:
	\newblock Hybrid loss for learning single-image-based hdr reconstruction.
	\newblock arXiv preprint:1812.07134 (2018)
	
	\bibitem{Wang19}
	Wang, C., Zhao, Y., Wang, R.:
	\newblock Deep inverse tone mapping for compressed images.
	\newblock IEEE Access \textbf{7} (2019)  74558--74569
	
	\bibitem{Soh19}
	Soh, J.W., Park, J.S., Cho, N.I.:
	\newblock Joint high dynamic range imaging and super-resolution from a single
	image.
	\newblock IEEE Access \textbf{7} (2019)  177427--177437
	
	\bibitem{Khan19}
	Khan, Z., Khanna, M., Raman, S.:
	\newblock Fhdr: Hdr image reconstruction from a single ldr image using feedback
	network.
	\newblock In: Proc. GlobalSIP, IEEE (2019)  1--5
	
	\bibitem{Marnerides21}
	Marnerides, D., Bashford-Rogers, T., Debattista, K.:
	\newblock Deep hdr hallucination for inverse tone mapping.
	\newblock Sensors \textbf{21} (2021)  4032
	
	\bibitem{Ye21}
	Ye, N., Huo, Y.,  et~al.:
	\newblock Single exposure high dynamic range image reconstruction based on deep
	dual-branch network.
	\newblock IEEE Access \textbf{9} (2021)  9610--9624
	
	\bibitem{Lee21}
	Lee, B.D., Sunwoo, M.H.:
	\newblock Hdr image reconstruction using segmented image learning.
	\newblock IEEE Access \textbf{9} (2021)  142729--142742
	
	\bibitem{Sharif21}
	A~Sharif, S., Naqvi, R.A.,  et~al.:
	\newblock A two-stage deep network for high dynamic range image reconstruction.
	\newblock In: Proc. CVPR. (2021)  550--559
	
	\bibitem{Zhang21}
	Zhang, Y., Ayd{\i}n, T.:
	\newblock Deep hdr estimation with generative detail reconstruction.
	\newblock Comput. Graph. Forum \textbf{40} (2021)  179--190
	
	\bibitem{Liu21}
	Liu, K., Cao, G.,  et~al.:
	\newblock Lightness modulated deep inverse tone mapping.
	\newblock arXiv preprint:2107.07907 (2021)
	
	\bibitem{Raipurkar21}
	Raipurkar, P., Pal, R., Raman, S.:
	\newblock Hdr-cgan: single ldr to hdr image translation using conditional gan.
	\newblock In: Proc. 12th Indian Conference on Computer Vision, Graphics and
	Image Processing. (2021)  1--9
	
	\bibitem{Yu21}
	Yu, H., Liu, W.,  et~al.:
	\newblock Luminance attentive networks for hdr image and panorama
	reconstruction.
	\newblock Comput. Graph. Forum (\textbf{40})  181--192
	
	\bibitem{Borrego22}
	Borrego-Carazo, J., Ozay, M.,  et~al.:
	\newblock A mixed quantization network for computationally efficient mobile
	inverse tone mapping.
	\newblock arXiv preprint:2203.06504 (2022)
	
	\bibitem{Wu22}
	Wu, G., Song, R.,  et~al.:
	\newblock Litmnet: A deep cnn for efficient hdr image reconstruction from a
	single ldr image.
	\newblock Pattern Recognition \textbf{127} (2022)  108620
	
	\bibitem{Endo17}
	Endo, Y., Kanamori, Y., Mitani, J.:
	\newblock Deep reverse tone mapping.
	\newblock ACM Trans. Graph. \textbf{36} (2017)  177--1
	
	\bibitem{Lee18}
	Lee, S., An, G.H.,  et~al.:
	\newblock Deep chain hdri: Reconstructing a high dynamic range image from a
	single low dynamic range image.
	\newblock IEEE Access \textbf{6} (2018)  49913--49924
	
	\bibitem{Lee18ECCV}
	Lee, S., An, G.H., Kang, S.J.:
	\newblock Deep recursive hdri: Inverse tone mapping using generative
	adversarial networks.
	\newblock In: Proc. ECCV. (2018)  596--611
	
	\bibitem{Jo21}
	Jo, S.Y., Lee, S.,  et~al.:
	\newblock Deep arbitrary hdri: Inverse tone mapping with controllable exposure
	changes.
	\newblock IEEE Trans. Multimedia (2021)
	
	\bibitem{Banterle22}
	Banterle, F., Marnerides, D.,  et~al.:
	\newblock Unsupervised hdr imaging: What can be learned from a single 8-bit
	video?
	\newblock arXiv preprint:2202.05522 (2022)
	
	\bibitem{Kinoshita19}
	Kinoshita, Y., Kiya, H.:
	\newblock Itm-net: Deep inverse tone mapping using novel loss function
	considering tone mapping operator.
	\newblock IEEE Access \textbf{7} (2019)  73555--73563
	
	\bibitem{Jang20}
	Jang, H.,  et~al.:
	\newblock Dynamic range expansion using cumulative histogram learning for high
	dynamic range image generation.
	\newblock IEEE Access \textbf{8} (2020)  38554--38567
	
	\bibitem{Cao21}
	Cao, G., Zhou, F.,  et~al.:
	\newblock A brightness-adaptive kernel prediction network for inverse tone
	mapping.
	\newblock Neurocomputing \textbf{464} (2021)  1--14
	
	\bibitem{Chambe22}
	Chambe, M., Kijak, E.,  et~al.:
	\newblock Hdr-lfnet: Inverse tone mapping using fusion network.
	\newblock hal preprint:03618267 (2022)
	
	\bibitem{AIM20ESR}
	Zhang, K.,  et~al.:
	\newblock Aim 2020 challenge on efficient super-resolution: Methods and
	results.
	\newblock In: Proc. ECCV. (2020)  5--40
	
	\bibitem{NTIRE22ESR}
	Li, Y.,  et~al.:
	\newblock Ntire 2022 challenge on efficient super-resolution: Methods and
	results.
	\newblock In: Proc. CVPR. (2022)  1062--1102
	
	\bibitem{MobileAI21FDN}
	Ignatov, A.,  et~al.:
	\newblock Fast camera image denoising on mobile gpus with deep learning, mobile
	ai 2021 challenge: Report.
	\newblock In: Proc. CVPR. (2021)  2515--2524
	
	\bibitem{NTIRE21}
	P{\'e}rez-Pellitero, E.,  et~al.:
	\newblock Ntire 2021 challenge on high dynamic range imaging: Dataset, methods
	and results.
	\newblock In: Proc. CVPR. (2021)  691--700
	
	\bibitem{FairchildHDR}
	Fairchild, M.D.:
	\newblock The hdr photographic survey.
	\newblock In: Color and imaging conference. Volume 2007. (2007)  233--238
	
	\bibitem{HdMHDR}
	Froehlich, J., Grandinetti, S.,  et~al.:
	\newblock Creating cinematic wide gamut hdr-video for the evaluation of tone
	mapping operators and hdr-displays.
	\newblock Digital photography X \textbf{9023} (2014)  279--288
	
	\bibitem{Liu21Retouching}
	Liu, Y., He, J.,  et~al.:
	\newblock Very lightweight photo retouching network with conditional sequential
	modulation.
	\newblock arXiv preprint:2104.06279 (2021)
	
	\bibitem{SFT}
	Wang, X., Yu, K.,  et~al.:
	\newblock Recovering realistic texture in image super-resolution by deep
	spatial feature transform.
	\newblock In: Proc. CVPR. (2018)  606--615
	
	\bibitem{PConv}
	Liu, G., Reda, F.A.,  et~al.:
	\newblock Image inpainting for irregular holes using partial convolutions.
	\newblock In: Proc. ECCV. (2018)  85--100
	
	\bibitem{DoubleJPEG}
	Jiang, J., Zhang, K., Timofte, R.:
	\newblock Towards flexible blind jpeg artifacts removal.
	\newblock In: Proc. ICCV. (2021)  4997--5006
	
	\bibitem{Hanji22}
	Hanji, P., Mantiuk, R., Eilertsen, G., Hajisharif, S., Unger, J.:
	\newblock Comparison of single image hdr reconstruction methods—the caveats
	of quality assessment.
	\newblock In: Proc. SIGGRAPH '22. (2022)  1--8
	
	\bibitem{VDP3}
	Wolski, K., Giunchi, D.,  et~al.:
	\newblock Dataset and metrics for predicting local visible differences.
	\newblock ACM Trans. Graph. \textbf{37} (2018)  1--14
	
	\bibitem{DnCNN-JPEG}
	Mathworks:
	\newblock Jpeg image deblocking using deep learning.
	\newblock
	(\url{mathworks.com/help/images/jpeg-image-deblocking-using-deep-learning.html})
	
	\bibitem{DnCNN}
	Zhang, K., Zuo, W.,  et~al.:
	\newblock Beyond a gaussian denoiser: Residual learning of deep cnn for image
	denoising.
	\newblock IEEE Trans. Img. Process. \textbf{26} (2017)  3142--3155
\end{thebibliography}
%

%
\end{document}